\documentclass{vgtc}                          

\usepackage{mathptmx}
\usepackage{graphicx}
\usepackage{times}
\usepackage{url}
\usepackage{flushend}

\onlineid{0}

\vgtccategory{Research}

\vgtcinsertpkg

\title{BodyDigitizer: An Open Source Photogrammetry-based 3D Body Scanner}

\author{Travis Gesslein
\and Daniel Scherer %
\and Jens Grubert\thanks{e-mail:jg@jensgrubert.de}}
\affiliation{\scriptsize Department of Electrical Engineering and Computer Science \\ 	Coburg University}

\vspace{2mm}
\teaser{

\centering
	\includegraphics[width=2.0\columnwidth]{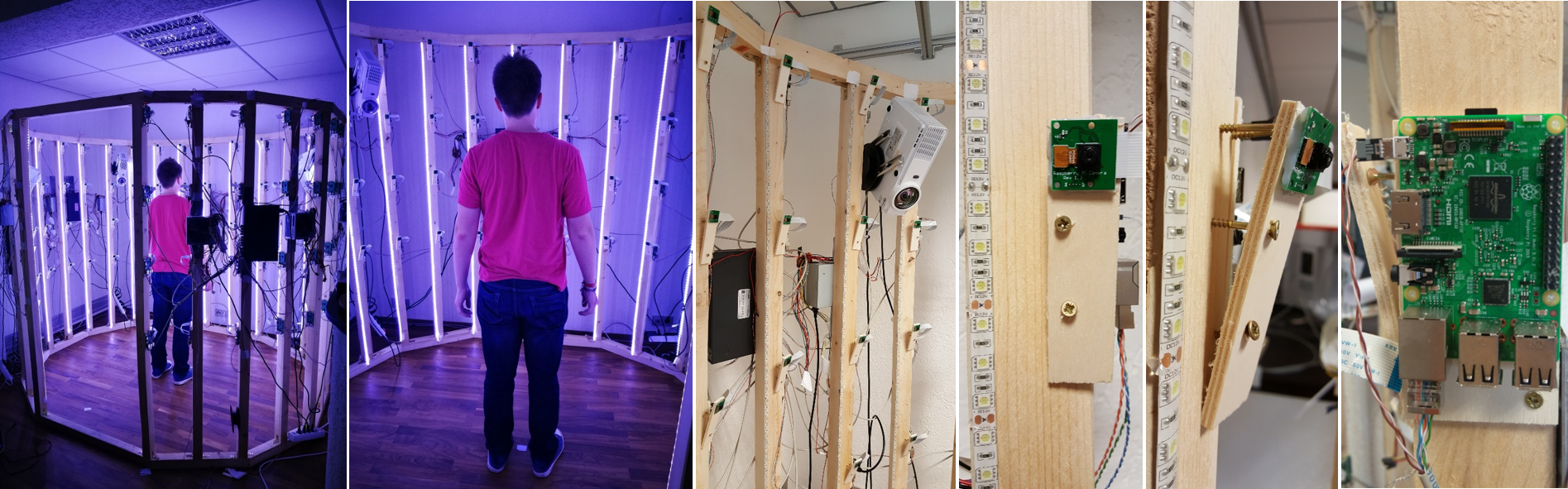}
\caption{From left to right: Outside view of the scanner with person and turned on LED strips; inside view; projector, power supplies and switches are mounted on the frame; detailed front view of camera mount; side view; front view of mounted Raspberry Pi.}
\label{fig:bodyscanner}
}

\abstract{
With the rising popularity of Augmented and Virtual Reality, there is a need for representing humans as virtual avatars in various application domains ranging from remote telepresence, games to medical applications. Besides explicitly modelling 3D avatars, sensing approaches that create person-specific avatars are becoming popular. However, affordable solutions typically suffer from a low visual quality and professional solution are often too expensive to be deployed in nonprofit projects.
We present an open-source project, BodyDigitizer, which aims at providing both build instructions and configuration software for a high-resolution photogrammetry-based 3D body scanner. Our system encompasses up to 96 Rasperry PI cameras, active LED lighting, a sturdy frame construction and open-source configuration software. 
The detailed build instruction and software are available at http://www.bodydigitizer.org.
} 

\CCScatlist{ 
 \CCScat{H.5.1}{Information Interfaces and Presentation (e.g. HCI)}{Multimedia Information Systems}{Artificial, augmented, and virtual realities}
}

\begin{document}


\firstsection{Introduction}

\maketitle


With the rise of consumer-oriented Augmented and Virtual Reality (AR/VR) systems and applications, there is a growing demand to represent person-specific 3D representations of humans. Applications that can strongly benefit from person-specific human 3D representations include remote telepresence \cite{Orts-Escolano:2016:HVT:2984511.2984517}, games and social VR \cite{fbs2017}, fitness applications \cite{naked2017} or medical applications \cite{molbert2017assessing}.

\begin{figure} [t]
	\begin{center}
		\includegraphics[width=0.9\columnwidth]{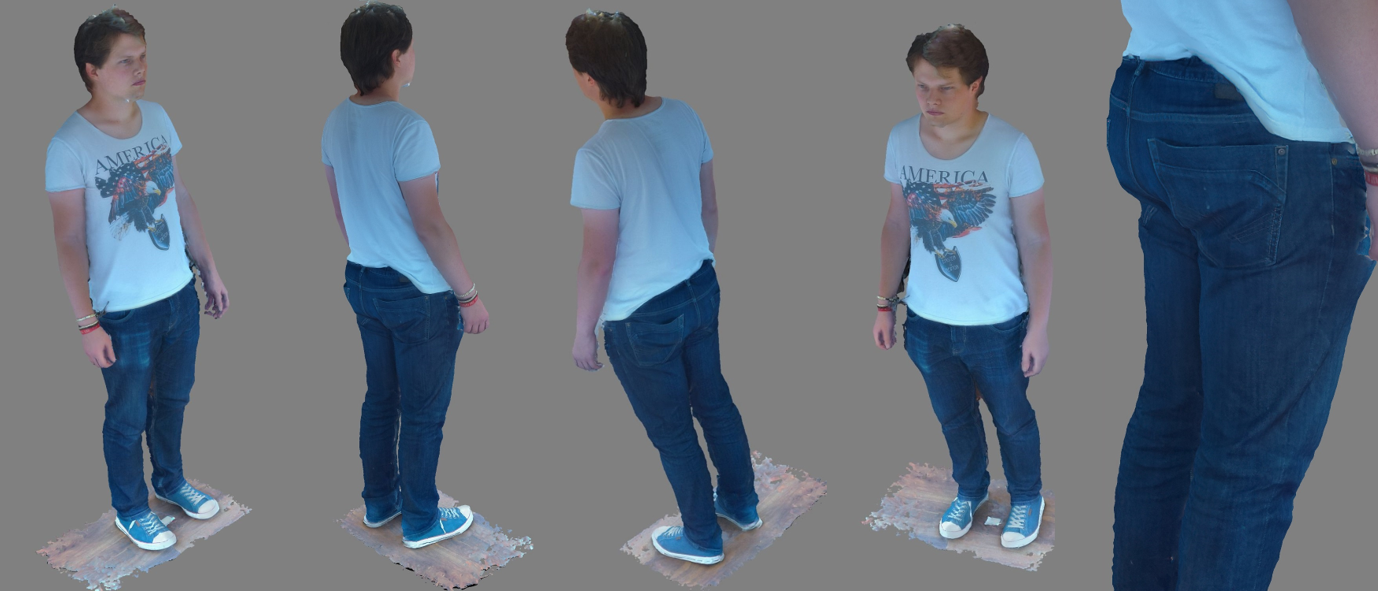}
		\caption{3D scan results from the proposed BodyDigitizer setup.}
		\label{fig:bodyscannerresultsfull}
	\end{center}
\end{figure}

There is a long history of capturing the appearance (and motion) of human bodies \cite{fuchs1994virtual},  
and open-source or low-cost approaches with commodity hardware suitable for full-body scans, emerged over the last years (e.g., \cite{tong2012scanning, scanoman2014, shapiro2014rapid, zhang2014quality, cui2012kinectavatar}). Single-sensor solutions, e.g. using commodity depth sensors like the Kinect \cite{cui2012kinectavatar} or Intel RealSense \cite{structure2017} result in comparatively low hardware costs but can suffer from lower spatial resolution compared to multi-view stereo setups \cite{smisek20133d} and require users to stand still for a prolonged period of time. Similar, utilizing single RGB-cameras for 3D reconstruction is popular in a wide variety of capturing tasks due to their wide-availability, e.g., city-scale 3D reconstructions \cite{agarwal2011building} or  hologram verification \cite{hartl2013mobile}, but can suffer from similar constraints as the single-depth camera-based approaches. In contrast, multi-camera rigs allow a fast capture of human bodies but can typically require a more elaborate physical setup. Within this paper, we focus on high-resolution reconstruction (see Figure \ref{fig:bodyscannerresultsfull}) with a short acquisition time using a multi-camera rig that relies on photogrammetry as 3D measurement principle. Specifically, we make available the build instructions as well as the acquisition software open-source.

\section{Related Work}

Commodity and open-source 3D body scanners have become popular over the last years.

Cui et al. proposed to scan a human body with a single Kinect \cite{cui20113d}, but the results were of low quality as the Cui's approach did not handle non-rigid movement or used color information. Weiss et al. \cite{weiss2011home} fitted a SCAPE model \cite{anguelov2005scape} to the 3D and image silhouette data from a Kinect, but failed to reproduce personalized details. Tong et al. used multiple stationary depth-sensors in combination with a turntable \cite{tong2012scanning}. Cui et al. simplified this setup to a single depth-sensor \cite{cui2012kinectavatar}. For this turntable-based approach, open-source build instructions have been made available \cite{scanoman2014}. Zhang et al. \cite{zhang2014quality} relaxed scanning requirements further and presented an approach that works with a handheld scanner. However, all those approaches result in long acquisition times, which can have negative impact on the scanning result (due to human motions) or be unacceptable for a given application scenario (e.g., scanning patients with bodily disorders for medical applications).

There also have been recent advances in real-time capture \cite{Orts-Escolano:2016:HVT:2984511.2984517}. However, also these state of the art approaches typically suffer from visible spatial and temporal artifacts, which might be undesirable for various applications. Further, for some applications the hard real-time constraints might not be necessary. 

Offline approaches typically result in higher quality models \cite{collet2015high} but can be cost intensive with commercial 3D scanners (e.g. Vitronic Vitus \cite{vitronic2017}) easily reaching a cost of more than 50.000 Euro). The closest work to ours is the PI3DSCAN project \cite{pi3d2017,garsthagenPi3D} and similar derivatives based on a Rasperry Pi infrastructure (e.g., \cite{straub2014development}). The PI3DSCAN project proposes a multi-camera setup for use with photogrammetry-based 3D reconstruction. While the PI3DSCAN project provides selected information regarding relevant hardware components, it also aims at selling products and services. Specifically, the project lacks concise assembly instructions or relevant open-source software for the management of the hardware components (e.g., scanning management software is sold for a yearly license of 750 Euro).  Hence, we aim at providing both concise built instructions as well as open-source acquisition software.

\section{3D Body Scanner}\label{sec:design}


Our 3D body scanner consists of a wooden frame,  electronic components built around Raspberry Pis and an acquisition software, which are described next. The total hardware cost of the scanner amounts to 9500 Euro, which is still substantially cheaper than commercial models, and can be decreased further with a lower number of cameras.

\subsection{Frame Construction and Mounts}

The frame was constructed to hold a flexible number of cameras with a minimum horizontal angle between cameras of ca. 13$^{\circ}$ (except for the entrance area, facing the back of the user). The shaped frame has the dimensions (width x depth x height) of 2.90m x 2.51m x 2.10m and a total number of 24 beams to ensure that mounted cameras with standard wide angle lenses could capture body parts with sufficient overlap. The total cost of the frame (wood, metal brackets, screws) was 150 Euro.

A challenge was to create affordable camera mounts. These camera mounts are required to position each camera module interactively when building the 3D scanner, and then hold them in place tightly. With 96 required mounts, even affordable consumer-oriented solutions like ball heads would quickly become expensive at scale (prices starting at 4 Euro per ball head). Also, 3D printing camera mounts would have been prohibitive, due to the amount of time required to print 96 holders. Instead, we opted for a simple solution consisting out of a wooden board on which the camera module was fixed with hot glue. This frame was then fixated with 3 screws (1 on the front, 2 on the back) on the underlying wooden bar, allowing to adjust the pose of the camera by screwing in or out the individual screws. The mounting procedure took on average 3 minutes and the cost was 12 Cent per mount.

For mounting Raspberry Pis to the frame, we used also used simple wooden boards. The Raspberry Pis were hot glued to the board and the board, subsequently, screwed to the wooden beams.

\begin{figure} [t]
	\begin{center}
		\includegraphics[width=1.0\columnwidth]{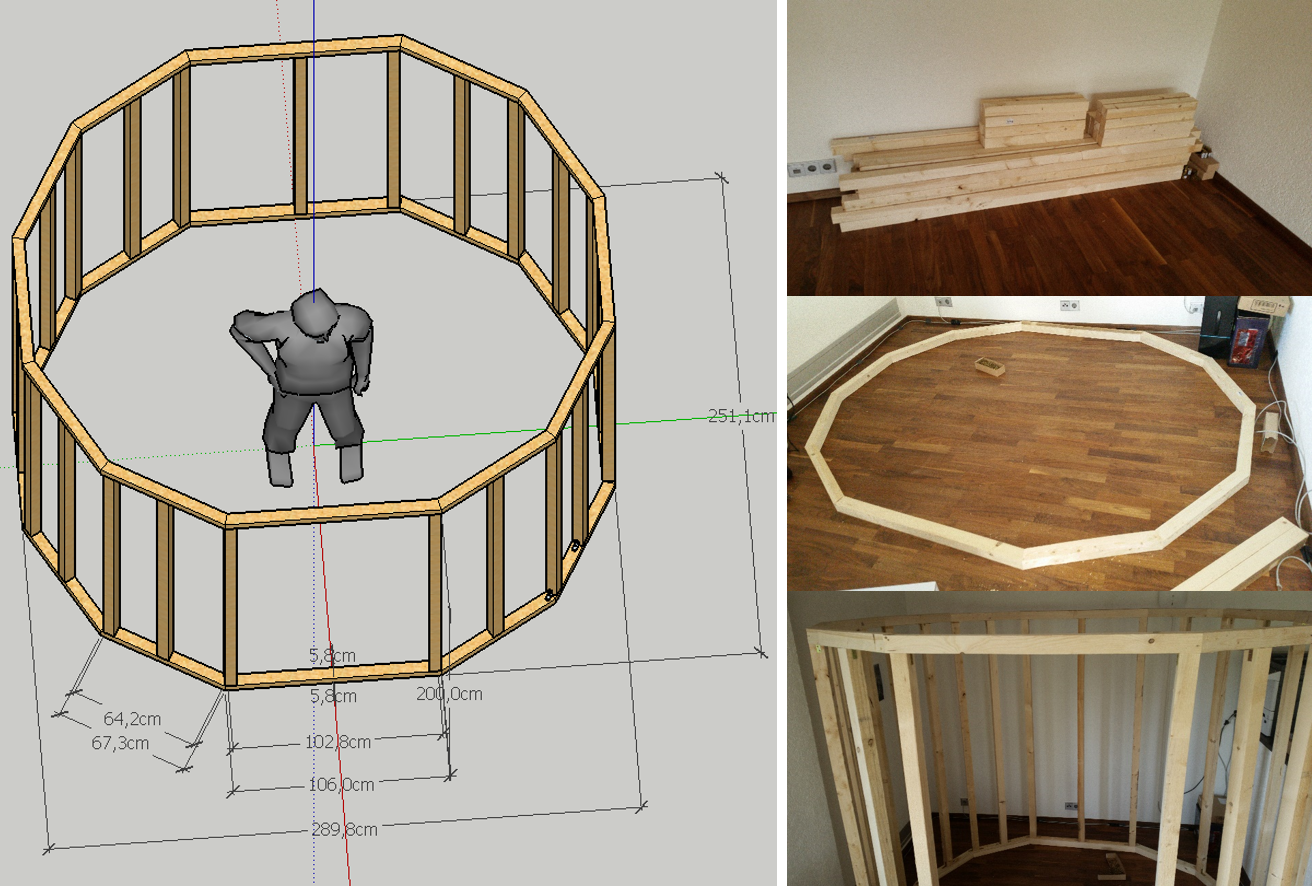}
		\caption{Left: 3D model of frame. Right: Assembly steps to final frame.}
		\label{fig:bs-construction}
	\end{center}
\end{figure}

\subsection{Electronic Components}

As we wanted to have flexible imaging pipeline enabling both high quality reconstruction and (in the future) live streaming, we chose 96 Raspberry Pis (Model 3) (4 per beam) in combination with Raspberry Pi Camera Module V1\footnote{https://www.raspberrypi.org/documentation/hardware/camera/} and a 8 GB micro-sd card. The total price of these components was 5.400 Euro. In addition, we wanted to have adjustable lighting and added 12V LED stripes with 60 LEDs and 1000-1300 lumen per meter as a primary light source (see Figure \ref{fig:bodyscanner}, for close-up view).

For power supply, we used ten 300 Watt PC power supplies with a 24 pole ATX connector. We cut off the connector and use the individual wires for 5V ( Raspberry Pi) and 12 V (LED stripes) power supply. As we wanted to minimize the number of cables used  (and the price per cable), we employed CAT 5 cables both for data transfer and power supply. Due to the small diameter individual wires, we limited the length of the power wires to \(80 cm\) in order to prevent voltage from dropping below the Pis specified minimum operating voltage of \( 4.75 V\). We arrived at this number by computing the voltage drop for an \(80 cm\) wire with diameter of \(0.27 mm^2\) and assumed maximum operating current of \(1.25 A\) for a single Pi, which results in a minimum voltage of \(4.8675 V\), leaving a subjectively chosen safety buffer of about \(0.1 A\) to protect against potential unaccounted factors in the practical setup. The maximum distance of \(80 cm\) also implied that the power supplies be mounted in the center of wooden beams serving power to two beams to the left and right. With longer cables, the Raspberry Pis would occasionally turn off during runtime as the potential drop would lead to a final voltage of \(4.75 V\) or below. 

To allow for LED lighting control, we use custom MOSFET boards, which can control four stripes via commands from a single Raspberry Pi (using its GPIO pins). Currently, we can control the light to be in 100\%, 50\% or 0\% level. In future, one could extend this setting to allow for shape from shading approaches using synchronized light patterns.

Finally, four short-throw projectors (Optoma GT 760) are used to project image patterns onto the human body to allow surface reconstruction in otherwise textureless regions. 




\subsection{Data Acquisition}

The acquisition software follows a client-server model and is based on node.js, chosen due to developer familiarity with the framework, as performance and stability requirements were small due to foreseeable lack of complexity in the software. The clients automatically try to (re-)connect to a known server IP at regular intervals, and locally have access to the 'raspistill' application used for taking camera images on Raspberry Pis. Using a simple command line interface the user can issue commands to all connected clients to take images or change projected image patterns. Once the image acquisition command has been issued, two sets of images are taken. First, an image set is created with projectors showing a black image (i.e. no projection) to be able to recreate the surface texture. Immediately after the first picture, a random dot pattern is projected onto the user and a second image is taken to allow for surface reconstruction. Then the images are automatically transferred to the PC. Transfer of a full 96 camera set of images with a resolution of 1920x1080 to the server in this manner takes 3-6 seconds, limited by the 1GB/s network card speed and read/write speed of SD cards used in the PI, which temporarily store images before sending. The Pis use Raspian [\cite{raspian}] as an operating system, which fulfills our stability and reliability requirements as no operating system-sourced outages were detected during runtime of the BodyDigitizer. Software maintenance of the 96 running systems is primarily achieved via the parallel-ssh and parallel-scp Linux command line tools \cite{pssh} \cite{pscp}, which enable parallel execution of SSH and SCP commands on an arbitrary number of target hosts.

For 3D reconstruction, we currently rely on a proprietary photo reconstruction software (Agisoft PhotoScan \cite{agisoft2017}), which automatically converts the reconstructed point cloud data into textured meshes. We plan to inspect alternative approaches 
in the future. For further use of the avatars, we currently employ a manual rigging and skinning process, but plan to investigate automated approaches \cite{feng2015avatar}.

\section{Limitations and Future Work}

\begin{figure} [t]
	\begin{center}
		\includegraphics[width=1\columnwidth]{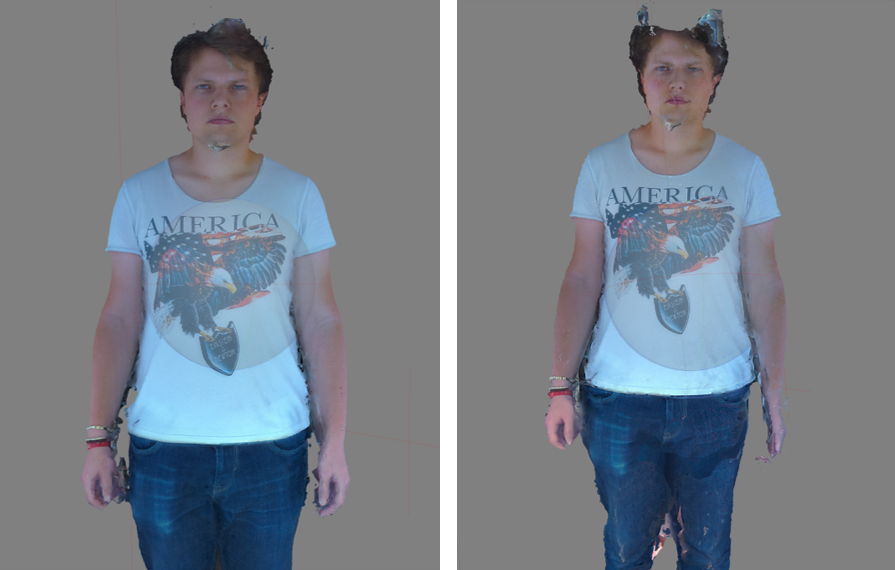}
		\caption{Left: Scan results using the full 96 camera setup. Right: Scan results using 48 cameras (every second camera removed in a chessboard pattern).}
		\label{fig:bodyscannerresultscompare}
	\end{center}
\end{figure}

While we were able to reduce the building cost to less than 10.000 Euro the required manual effort is still substantial, requiring approximately two person months to build such a system. Depending on the required model fidelity, a smaller number of cameras (the current cost for a single Raspberry Pi 3 + camera + sd card is approximately 56 Euro) might be sufficient. For example, in Figure \ref{fig:bodyscannerresultscompare}, right, we used 48 cameras, in comparison to the full setup with 96 cameras in Figure \ref{fig:bodyscannerresultscompare}, left. Also, depending on the use case (e.g., expected clothing), the four projectors (with a total cost of 2.000 Euro) might be optional. 

While we have completed an acquisition software, we still rely on commercial solutions for the actual 3D reconstruction of the models. Further, ideally, the model scanning, conversion, rigging, skinning and provisioning into the VR scene should be fully automated, in order to a support swift working procedures when preparing and conducting experiments. 

In future work, we want to address these issues by providing a fully automated and flexible body scanning and streaming pipeline. Specifically, we aim at the inclusion of image-based visual hull algorithms \cite{matusik2000image} employing the already installed Raspberry Pi cameras or the inclusion of multiple commodity depth cameras \cite{kainz2012omnikinect} for streaming purposes. 

Further, the setup can currently not reconstruct virtual avatars in real-time, limited by both image transferal as well as model generation speed in Agisoft Photoscan. We want to investigate real-time applications in the domain of pervasive AR \cite{grubert2017towards} by possibly reducing image resolution and investigating different algorithms for model generation. Also, cameras are currently calibrated as part of the algorithm pipeline in Photoscan. But since camera configuration is static, we want to investigate ways to pre-calibrate cameras and reuse the resulting data to skip those algorithm steps.

\section{Conclusion}
We have presented an open-source project, BodyDigitizer, which aims at providing both built instructions and configuration software for a high-resolution photogrammetry-based 3D body scanner. Our system encompasses up to 96 Rasperry PI cameras, active LED lighting, a sturdy frame construction and open-source configuration software. 
The detailed build instruction and software are available at http://www.bodydigitizer.org.

\bibliographystyle{abbrv}
\bibliography{bodydigitizer}

\begin{thebibliography}{10}

\bibitem{vitronic2017}
3d body scanner - vitronic - the machine vision peopl.
\newblock
  \url{https://www.vitronic.com/industrial-and-logistics-automation/sectors/3d-body-scanner.html}.
\newblock Accessed: 2017-09-11.

\bibitem{agisoft2017}
Agisoft photoscan.
\newblock \url{http://www.agisoft.com/}.
\newblock Accessed: 2017-09-11.

\bibitem{fbs2017}
Facebook spaces.
\newblock \url{https://www.facebook.com/spaces}.
\newblock Accessed: 2017-09-11.

\bibitem{naked2017}
Naked fit.
\newblock \url{https://naked.fit/}.
\newblock Accessed: 2017-09-11.

\bibitem{pscp}
Parallel scp.
\newblock \url{https://linux.die.net/man/1/pscp}.
\newblock Accessed: 2017-10-13.

\bibitem{pssh}
Parallel ssh.
\newblock \url{https://linux.die.net/man/1/pssh}.
\newblock Accessed: 2017-10-13.

\bibitem{pi3d2017}
Pi3d 3d body scanner.
\newblock \url{http://www.pi3dscan.com/}.
\newblock Accessed: 2017-09-11.

\bibitem{raspian}
Raspian operating system.
\newblock \url{https://www.raspbian.org/}.
\newblock Accessed: 2017-10-13.

\bibitem{scanoman2014}
Scanoman - open source full-body 3d scanner.
\newblock \url{http://3dmaker4u.com/scanoman-project.html}.
\newblock Accessed: 2017-09-11.

\bibitem{structure2017}
Structure sensor - 3d scanning, augmented reality, and more for mobile devices.
\newblock \url{http://structure.io}.
\newblock Accessed: 2017-09-11.

\bibitem{agarwal2011building}
S.~Agarwal, Y.~Furukawa, N.~Snavely, I.~Simon, B.~Curless, S.~M. Seitz, and
  R.~Szeliski.
\newblock Building rome in a day.
\newblock {\em Communications of the ACM}, 54(10):105--112, 2011.

\bibitem{anguelov2005scape}
D.~Anguelov, P.~Srinivasan, D.~Koller, S.~Thrun, J.~Rodgers, and J.~Davis.
\newblock Scape: shape completion and animation of people.
\newblock In {\em ACM Transactions on Graphics (TOG)}, volume~24, pages
  408--416. ACM, 2005.

\bibitem{collet2015high}
A.~Collet, M.~Chuang, P.~Sweeney, D.~Gillett, D.~Evseev, D.~Calabrese,
  H.~Hoppe, A.~Kirk, and S.~Sullivan.
\newblock High-quality streamable free-viewpoint video.
\newblock {\em ACM Transactions on Graphics (TOG)}, 34(4):69, 2015.

\bibitem{cui2012kinectavatar}
Y.~Cui, W.~Chang, T.~N{\"o}ll, and D.~Stricker.
\newblock Kinectavatar: fully automatic body capture using a single kinect.
\newblock In {\em Asian Conference on Computer Vision}, pages 133--147.
  Springer, 2012.

\bibitem{cui20113d}
Y.~Cui and D.~Stricker.
\newblock 3d shape scanning with a kinect.
\newblock In {\em ACM SIGGRAPH 2011 Posters}, page~57. ACM, 2011.

\bibitem{feng2015avatar}
A.~Feng, D.~Casas, and A.~Shapiro.
\newblock Avatar reshaping and automatic rigging using a deformable model.
\newblock In {\em Proceedings of the 8th ACM SIGGRAPH Conference on Motion in
  Games}, pages 57--64. ACM, 2015.

\bibitem{fuchs1994virtual}
H.~Fuchs, G.~Bishop, K.~Arthur, L.~McMillan, R.~Bajcsy, S.~Lee, H.~Farid, and
  T.~Kanade.
\newblock Virtual space teleconferencing using a sea of cameras.
\newblock In {\em Proc. First International Conference on Medical Robotics and
  Computer Assisted Surgery}, volume~26, 1994.

\bibitem{garsthagenPi3D}
R.~Garsthagen.
\newblock An open source, low-cost, multi camera full-body 3d scanner.
\newblock In {\em 5th International Conference on 3D Body Scanning
  Technologies}, 2014.

\bibitem{grubert2017towards}
J.~Grubert, T.~Langlotz, S.~Zollmann, and H.~Regenbrecht.
\newblock Towards pervasive augmented reality: Context-awareness in augmented
  reality.
\newblock {\em IEEE transactions on visualization and computer graphics},
  23(6):1706--1724, 2017.

\bibitem{hartl2013mobile}
A.~Hartl, J.~Grubert, D.~Schmalstieg, and G.~Reitmayr.
\newblock Mobile interactive hologram verification.
\newblock In {\em Mixed and Augmented Reality (ISMAR), 2013 IEEE International
  Symposium on}, pages 75--82. IEEE, 2013.

\bibitem{kainz2012omnikinect}
B.~Kainz, S.~Hauswiesner, G.~Reitmayr, M.~Steinberger, R.~Grasset, L.~Gruber,
  E.~Veas, D.~Kalkofen, H.~Seichter, and D.~Schmalstieg.
\newblock Omnikinect: real-time dense volumetric data acquisition and
  applications.
\newblock In {\em Proceedings of the 18th ACM Symposium on Virtual Reality
  Software and Rechnology}, pages 25--32. ACM, 2012.

\bibitem{matusik2000image}
W.~Matusik, C.~Buehler, R.~Raskar, S.~J. Gortler, and L.~McMillan.
\newblock Image-based visual hulls.
\newblock In {\em Proceedings of the 27th Annual Conference on Computer
  graphics and Interactive Techniques}, pages 369--374. ACM
  Press/Addison-Wesley Publishing Co., 2000.

\bibitem{molbert2017assessing}
S.~M{\"o}lbert, A.~Thaler, B.~Mohler, S.~Streuber, J.~Romero, M.~Black,
  S.~Zipfel, H.-O. Karnath, and K.~Giel.
\newblock Assessing body image in anorexia nervosa using biometric self-avatars
  in virtual reality: Attitudinal components rather than visual body size
  estimation are distorted.
\newblock {\em Psychological Medicine}, pages 1--12, 2017.

\bibitem{Orts-Escolano:2016:HVT:2984511.2984517}
S.~Orts-Escolano, C.~Rhemann, S.~Fanello, W.~Chang, A.~Kowdle, Y.~Degtyarev,
  D.~Kim, P.~L. Davidson, S.~Khamis, M.~Dou, V.~Tankovich, C.~Loop, Q.~Cai,
  P.~A. Chou, S.~Mennicken, J.~Valentin, V.~Pradeep, S.~Wang, S.~B. Kang,
  P.~Kohli, Y.~Lutchyn, C.~Keskin, and S.~Izadi.
\newblock Holoportation: Virtual 3d teleportation in real-time.
\newblock In {\em Proceedings of the 29th Annual Symposium on User Interface
  Software and Technology}, UIST '16, pages 741--754, New York, NY, USA, 2016.
  ACM.

\bibitem{shapiro2014rapid}
A.~Shapiro, A.~Feng, R.~Wang, H.~Li, M.~Bolas, G.~Medioni, and E.~Suma.
\newblock Rapid avatar capture and simulation using commodity depth sensors.
\newblock {\em Computer Animation and Virtual Worlds}, 25(3-4):201--211, 2014.

\bibitem{smisek20133d}
J.~Smisek, M.~Jancosek, and T.~Pajdla.
\newblock 3d with kinect.
\newblock In {\em Consumer depth cameras for computer vision}, pages 3--25.
  Springer, 2013.

\bibitem{straub2014development}
J.~Straub and S.~Kerlin.
\newblock Development of a large, low-cost, instant 3d scanner.
\newblock {\em Technologies}, 2(2):76--95, 2014.

\bibitem{tong2012scanning}
J.~Tong, J.~Zhou, L.~Liu, Z.~Pan, and H.~Yan.
\newblock Scanning 3d full human bodies using kinects.
\newblock {\em IEEE Transactions on Visualization and Computer Graphics},
  18(4):643--650, 2012.

\bibitem{weiss2011home}
A.~Weiss, D.~Hirshberg, and M.~J. Black.
\newblock Home 3d body scans from noisy image and range data.
\newblock In {\em Computer Vision (ICCV), 2011 IEEE International Conference
  on}, pages 1951--1958. IEEE, 2011.

\bibitem{zhang2014quality}
Q.~Zhang, B.~Fu, M.~Ye, and R.~Yang.
\newblock Quality dynamic human body modeling using a single low-cost depth
  camera.
\newblock In {\em Proceedings of the IEEE Conference on Computer Vision and
  Pattern Recognition}, pages 676--683, 2014.

\end{thebibliography}
\end{document}